# Vastextures: Vast repository of textures and PBR materials extracted from real-world images using unsupervised methods

Sagi Eppel[1]

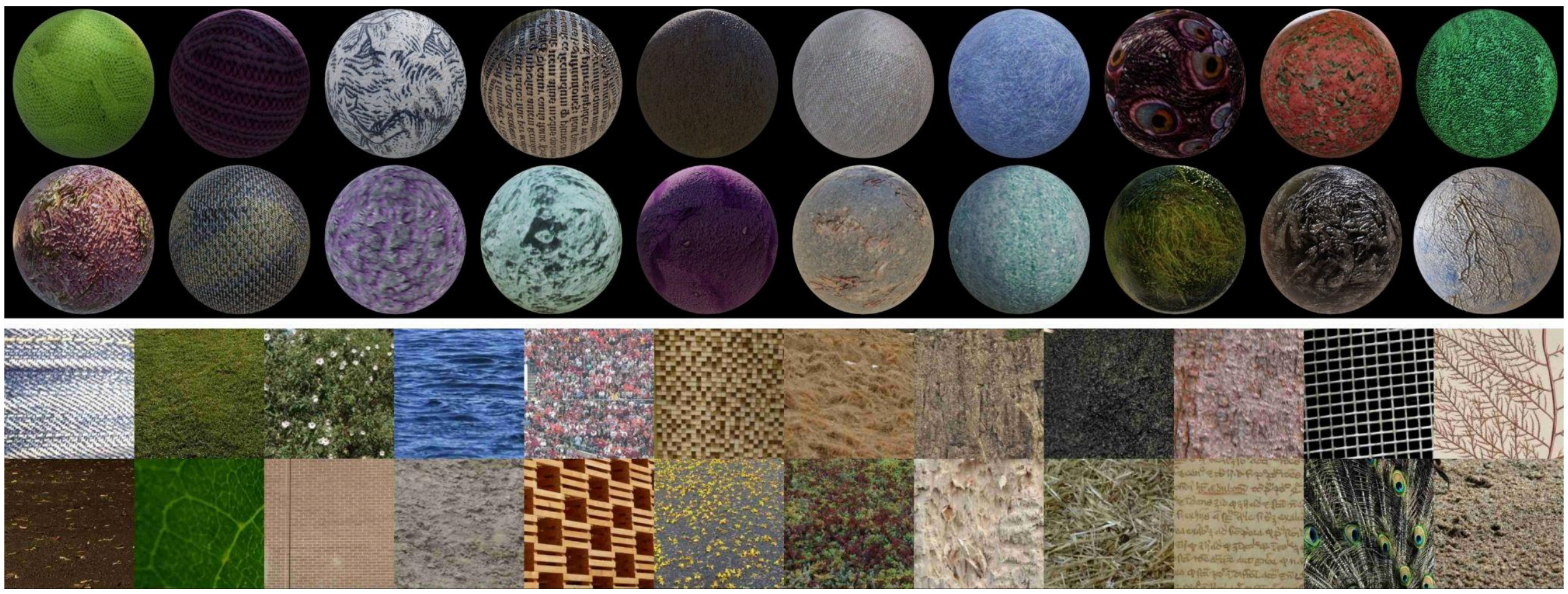

## Abstract[1]

Vastextures is a vast repository of 500,000 textures and PBR materials extracted from real-world images using an unsupervised process. The extracted materials and textures are extremely diverse and cover a vast range of real-world patterns, but at the same time less refined compared to existing repositories. The repository is composed of 2D textures cropped from natural images and SVBRDF/PBR materials generated from these textures. Textures and PBR materials are essential for CGI. Existing materials repositories focus on games, animation, and arts, that demand a limited amount of high-quality assets. However, virtual worlds and synthetic data are becoming increasingly important for training A.I systems for computer vision. This application demands a huge amount of diverse assets but at the same time less affected by noisy and unrefined assets. Vastexture aims to address this need by creating a free, huge, and diverse assets repository that covers as many real-world materials as possible. The materials are automatically extracted from natural images in two steps: 1) Automatically scanning a giant amount of images to identify and crop regions with uniform textures. This is done by splitting the image into a grid of cells and identifying regions in which all of the cells share a similar statistical distribution.
2) Extracting the properties of the PBR material from the cropped texture. This is done by randomly guessing every correlation between the properties of the texture image and the properties of the PBR material. The resulting PBR materials exhibit a vast amount of real-world patterns as well as unexpected emergent properties. Neutral nets trained on this repository outperformed nets trained using handcrafted assets. 500,000 textures and PBR materials have been made available at 1, 2, 3, 4.

1 The Vastexture repository was published as part of "**Infusing Synthetic Data with Real-World Patterns for Zero-Shot Material State Segmentation**"[1]. This work gives a more focused discussion of this dataset.

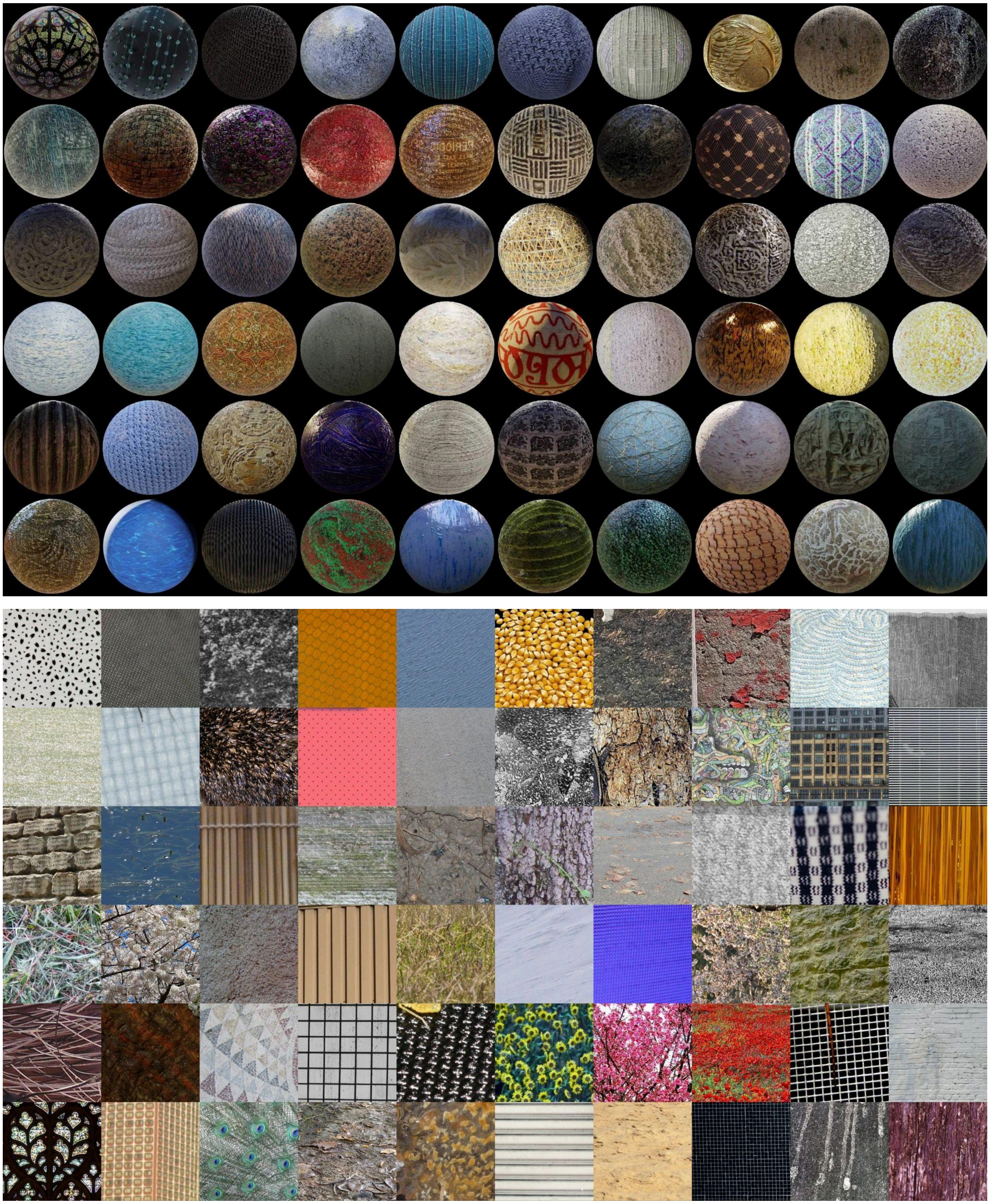

**Figure 1) Samples PBR materials (top) and 2D textures (bottom) from the Vastextures repository. 500,000 textures and SVBRDF/PBR materials have been made free and publicly available.**

# 1.Introduction

Textures and physics-based rendering (PBR) materials are crucial for representing the visual appearance of materials in computer-generated images (CGI) for animation, games, and other arts[2-14]. Textures and PBR materials also become increasingly important for generating virtual worlds and synthetic data for training artificial intelligence systems[15-24]. Materials are usually represented either as 2D textures for 2D scenes or using physics-based rendering (PBR) materials (also called SVBRDF materials) which describe the distribution of material properties as a set of maps. Each map describes the distribution of one property on the material surface[25,26]. Existing textures and PBR materials repositories[2-14] are mostly created using a handcrafted approach in which artists identify interesting textures in images and use intuition to "guess" the physical properties of the material from the RGB image[27-30]. An alternative approach uses A.I (trained on handcrafted PBRs) to generate new PBR materials from input text or images[31-41]. Both handcrafted and AI-based approaches are focused on generating assets for CGI artists that demand highly refined visually appealing assets. However, handcraft methods are limited by the quantity of assets that can be created[27-30]. While A.I is limited by the diversity of the data on which it was trained, which is mostly composed of handcrafted assets[31-41]. As such they have limited diversity and are not well suited for generating the diverse and huge amount of data needed for training modern neural nets. Vastexture aims to fill this gap by creating a giant and highly diverse repository that is extracted from natural images with minimal restrictions. The core idea is to cast a wide net, capturing as many textures and real-world patterns even at the cost of capturing unrelated patterns and more noisy assets in the process. The repository is composed of two parts: 1) 2D textures extracted from real-world images (Figure 1 bottom), and 2) PBR materials generated from these textures (Figure 1 top).

**2D texture repositories and datasets:** have been available mostly for artists and academic research. These are usually limited to a set of a few hundred to a few thousand images of textures divided into a few classes[2-14,32,42-52]. In contrast, the Vastexture contains 250,000 textures without being limited to given categories or domains.

**PBR materials repositories:** Existing PBR material repositories are composed of a few hundred to a few thousand high-quality manually or made materials[3,4.9-14]. The Vastexture repository does not aim to compete with these repositories but rather to supplement them in their main limitation: the limited number of assets and their diversity. The Vastextures contain 300,000 PBR materials (not including mixes), each extracted from a unique texture image.

## 1. General approach

Extracting textures and PBR materials from images is done in two steps:

1. Identifying and cropping image regions with uniform textures, which often imply uniform material.
2. Using the cropped texture to guess the physical properties of the SVBRDF/PBR material it represents.

# 3. Identifying and extracting uniform textures from images

Extracting textures from images is mostly done either manually or using A.I. Both approaches can generate high-quality assets for CGI art. However, manual extraction is limited in quantities. While A.I-based approaches are limited by the diversity of the assets they were trained on[31-41].

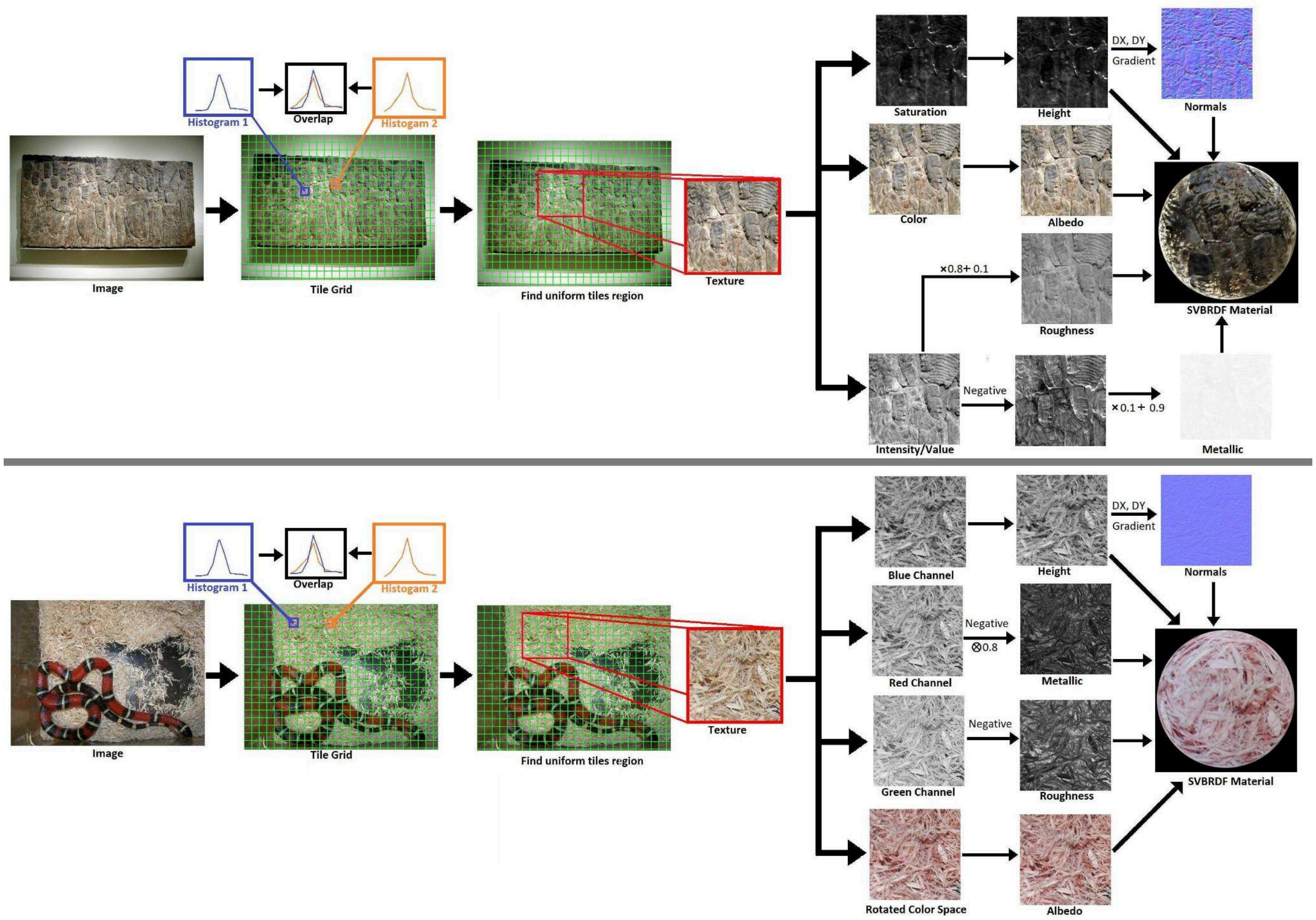

**Figure 2) Material extraction methods in two examples: Pick an image and divide it into a grid of square cells. For every grid cell, extract the distribution of features (color, gradients). Identify image regions for which all cells have similar distributions as a uniform texture. d,e) Pick random channels from the extracted texture image, augment them, and use the resulting maps as property maps (roughness, metallic, height, etc.) for the PBR material.**

## 3.1. Extracting textures from images using statistical similarity

An image texture can be defined as an area with some uniform distribution of patterns. Hence, we can find textures by measuring the distribution of patterns across the image and search for square regions in which this distribution is uniform. More accurately we can split the image into a grid and identify regions in which all the cells share a similar distribution of color and gradients as having uniform textures (Figure 2). Methods for extracting textures by statistical similarity[53-58] have been heavily explored since the 1970s, but were mostly discarded in favor of A.I and machine learning tools. This is largely because the extracted textures are noisy and include not only materials but any region with repeating patterns like crowds of people, waves at sea, or uniform blocks of buildings (Figures 1,4). In addition, these methods are sensitive to uneven illumination and will not extract regions with uniform materials but uneven light and shadows. As such statistical methods are far less useful for the high-quality assets needed for CGI artists. In contrast, neural nets like GPT[59] and CLIP[60,61] prove that A.I can learn from noisy data and achieve high levels of world understanding. On the other hand, nets trained on clean but narrow domains of data performed poorly when applied to general real-world tasks[60,62]. Hence, the highly general but noisy nature of patterns extracted by statistical methods is very well suited for training neural nets. While the method might miss significant amounts of textures due to non-uniform illuminations, we can expect that for a large enough set of images, any important texture will appear in many settings and at least in some cases under uniform light where it will be easily identified and extracted. Hence, by scanning a large enough set of images we are likely to extract any major material or texture. Samples of extracted textures are shown in Figures 1,4. It can be seen that this approach captures a wide range of materials but at the same time captures many unrelated regular patterns like waves at sea, dense vegetation, crowds of people or tiled architectures.

### 3.3. Detailed implementation

The approach for texture extraction (Figure 2) uses the following steps :

1. Resize and split the image into a grid of square cells of 40x40 to 75x75 pixels each (75x75 tiles were used for larger images in the segment anything dataset[63]).
2. Each cell collects the distribution/histogram of feature values in the cell (features are the R,G,B colors and their gradients).
3. Identify regions in which all cells share a similar distribution of features as having uniform textures. In other words, regions of 7x7 cells or more in which every pair of cells share a similar distribution of color and gradients are considered to belong to the same texture and are extracted.
4. The similarity between two cells is measured by the Jensen-Shannon distance of their colors and gradients distribution. Colors taken as R,G,B image channels and gradients are the vertical and horizontal differences between neighboring pixels (sobel filter).
5. Filter regions with too uniform a distribution of color, as these just represent smooth featureless regions.

Note that a lot of work has been done in this field[53-62], and the goal here is not to take the most sophisticated approach but rather the simplest. Images were taken from the segment anything repository[63] and the open images repository[64,65]. Parameters like image resizing, cell size, minimal texture size, and threshold, were all manually tuned for each repository by visual inspection.

## 4. Extracting SVBRDF/PBR materials properties from texture images

In order to simulate materials in 3D environments it's not enough to have a picture of the texture. To predict how the material will appear from different angles and illuminations it is necessary to know various physical properties like albedo, roughness, transparency, and normals, and how they vary across the material surface[25-30]. Hence, we need to take the RGB texture and extract a map for each material property from it (Figure 2). There is no deterministic way to extract material properties from a simple RGB image. All options involve guessing in one way or another. Guessing can be based on experience and intuition for handcrafted assets[27-30], or statistical learning for the A.I methods[31-41]. Were again the A.I learn by seeing a large number of examples handcrafted by humans[32]. Hence, A.I is limited by the diversity of training examples, while humans are limited by the number of assets they can create.

## 4.1. Casting a wide net: Guessing every correlation

A.I and manual approaches try to guess the most likely properties of the material in the image. In contrast,our approach is to guess every possible correlation between simple image properties (RGB,HSV…) and the PBR material property maps (Rougnes, height, metallic…). More accurately to assume that the material properties are correlated with the texture image properties in a simple way, and therefore can be derived from them. For example, more reflective regions are brighter or cracks are darker. However, the exact way in which the image properties are correlated with the material properties is unknown. The solution is not to try to find the most statistically likely correlation but instead to guess every correlation between image property and material property (Figure 2). Hence, for every PBR material property (roughness, metallic, height, transmission...) we randomly pick one of the image property maps (R,G,B, H,S,V…). For example, we may pick the brightness channel of the image randomly, scale and augment it and then use it as the roughness map of the PBR material (Figure 2). Or pick the red channel of the image and use it with some augmentation as the height map. For the normal map, we take the gradient of the height map. For all maps except normals we also randomly use uniform random values, and soft and hard thresholding in some cases.

## 4.2. Explanation and justification

The reasoning beyond the above approach is that the resulting PBR materials will end up using all patterns in the texture image, and will use them for each and every material property. Hence, we get maximum variability and maximum utilization of the patterns in the image, with minimum assumptions. As a result all the information we can extract from the image will be embedded in all physical properties a PBR can simulate (for a large enough set). Since the textures themselves are extracted from a vast range of images in an unsupervised way and therefore contain almost unlimited amounts of real-world patterns. This means that the resulting PBRs will likely represent any important or common texture and pattern in the world, in any of its physical properties (Again the assumption here is that important materials patterns will appear in a wide range of images and will be extracted at least in some of them.)

This approach is very different from A.I and manual approaches that try to guess the single most likely correlation based on intuition or past examples, which lead to more accurate but also narrow distribution materials. The Vastexture PBRs are far less likely to accurately represent the properties of the material in the image but will cover a far broader range of domains, and are far more likely to embed the full possibilities and complexity of patterns in the world.

## 4.3. Detailed implementation

The generating of materials follows the following steps (Figure 2.right):

1. For each property of the PBR material (albedo, roughness, metallic, height, transmission..) randomly choose one or more properties mapped in the image. The property maps include the Red, Green, and Blue channels and the Hue, Values, and Saturation channels.
2. Randomly Augment this map by using rotating color space, Scaling by a random value (multiply), adding random value, taking the negative of the map (1 - map), or soft or hard thresholding, and color ramp. Apply this augmentation randomly.
3. Use the augmented map (1,2) to represent the property map of the PBR.
4. For all properties other than normal and color in a fraction of cases give the property a random uniform value.
5. For normal maps use the gradient of the height map, with a random scaling.
6. For base color/albedo use the RGB of the image. In a small fraction of cases use augmentation like rotating color space.

## 4.4. Resulting PBRs materials

Samples of the resulting PBRs are shown in Figures 1,3 It can be seen that these materials contain a vast amount of real-world material textures, as well as things like waves, branches, feathers, and many others. Some of the generated PBR seems to be a good representation of the materials in the image from which it was generated. However, many of the generated materials show emerging, unique properties not appearing in images, but nonetheless representing possible materials. For example, patterns of tree branches can be mapped into lower regions' height maps, creating crack-like structures. When this branching structure is mapped to a bump map it gives vein-like patterns. These emergent properties are similar procedural generation techniques[66-68]. However, procedural textures are limited to a set of predefined building blocks, while for Vastexture the building blocks are by themself unlimited natural patterns.

# 5. Using Vastexture materials for training A.I systems.

To examine the validity of this repository for training neural nets. We use the MatSeg dataset for zero-shot segmentation of materials[1]. We re-created this dataset twice. Once using handcrafted assets that encompass almost all the free publicly available PBRs, and second using the Vastexture PBRs. The results show that the net trained on data constructed using the Vastexture achieved 92% accuracy compared to 89% for the net trained on handcrafted assets.

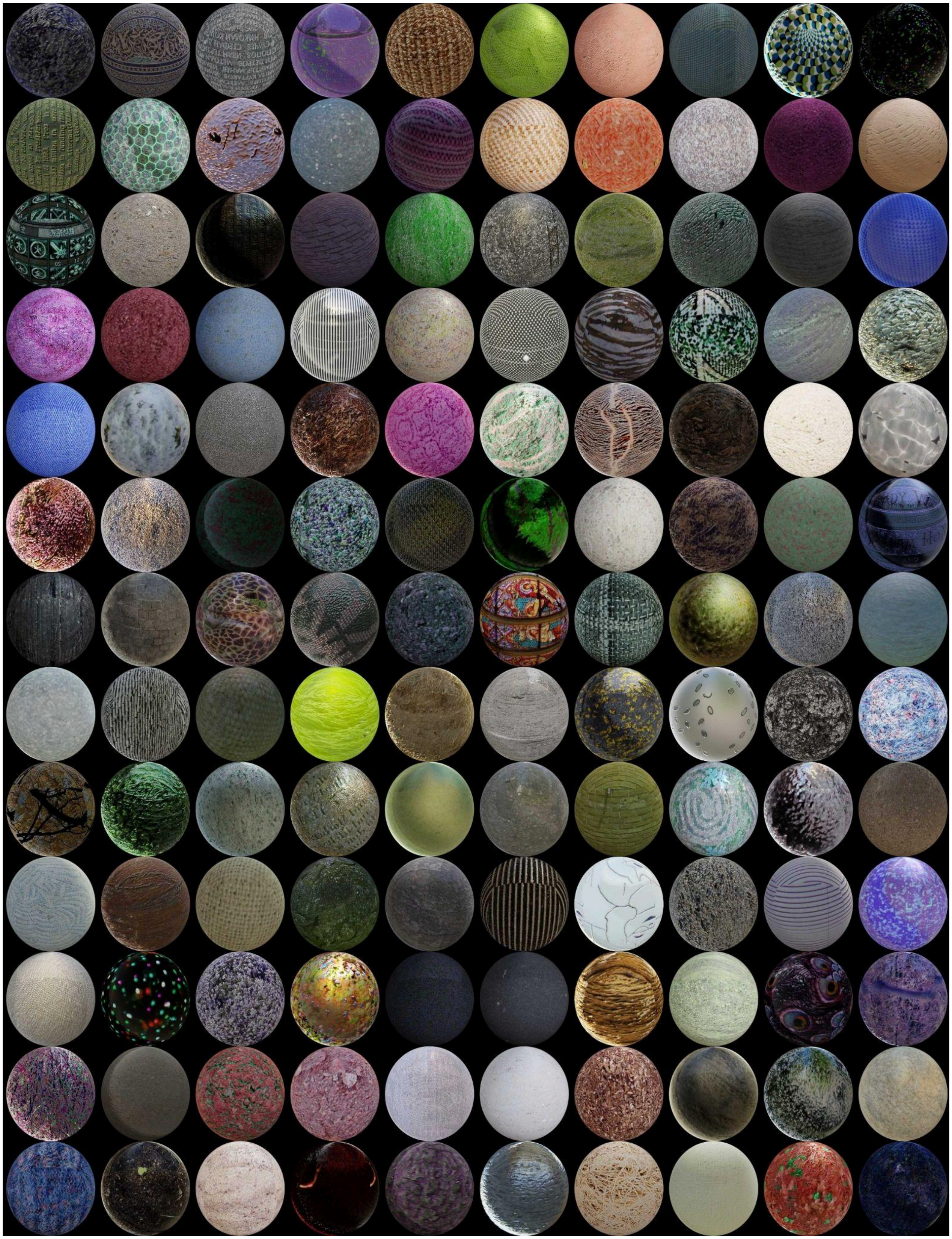

**Figure 3) More samples of Vastexture PBR materials.**

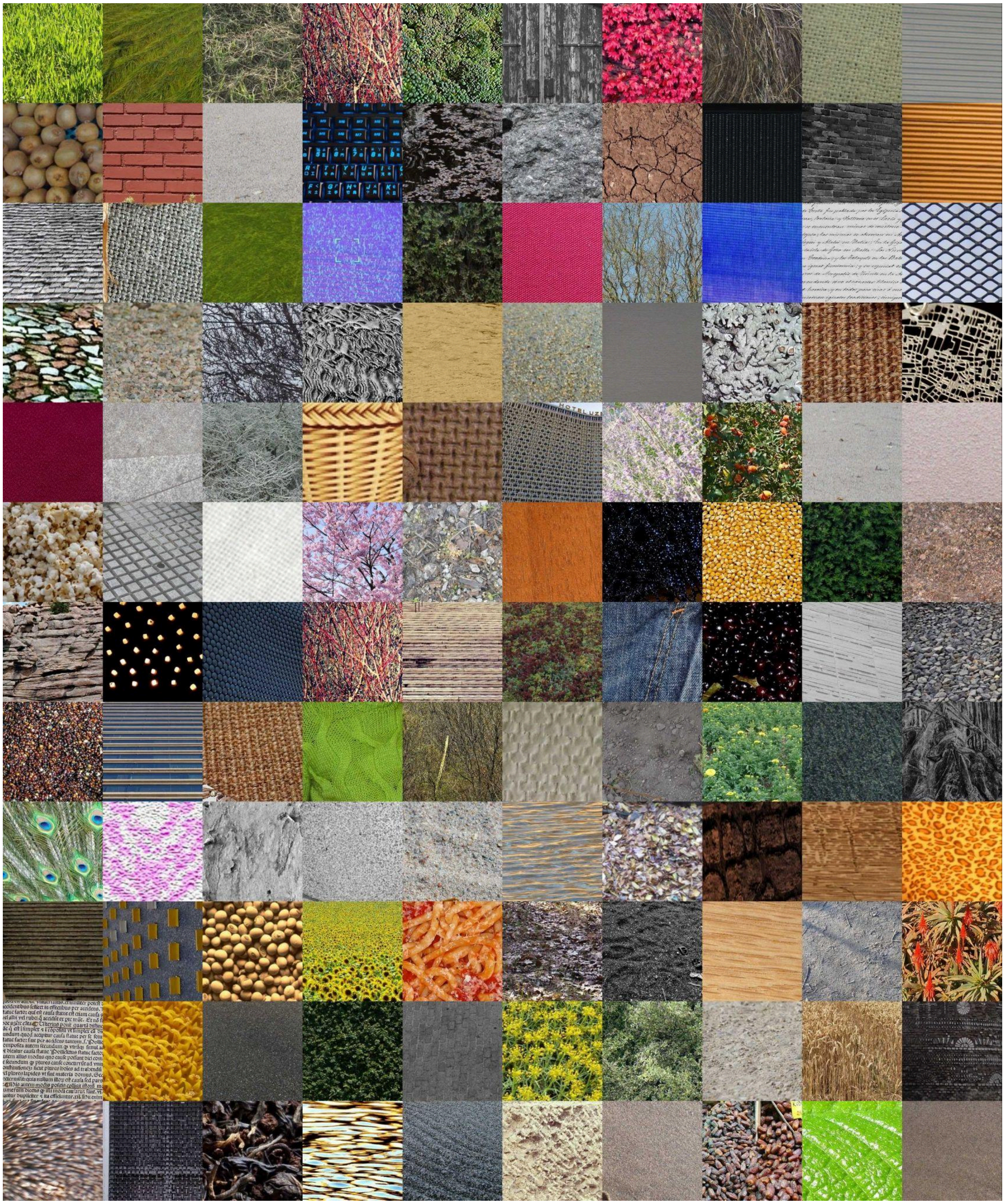

**Figure 4) More samples of extracted textures (textures extracted from the open images dataset).**

# Appendix

## A7. Mixing PBR materials,

Another way to increase the complexity of PBR materials is by mixing two or more materials. Similar to past work this is done by mixing the property maps of the two textures using a random mixing ratio.  For each property in the PBR material, the weighted average of the corresponding maps in the two materials. The mixing ratio was determined randomly for each property separately (different value for each property), or single mixing value for all the properties.

## A8. Seamless textures and tilability

Textures in the real world are rarely seamless or tileable, meaning they cannot be tiled periodically without creating visible seamlines (Figure 5 left) as a result the extracted textures in Vastextures are in most cases not tileable by default. This limits their usefulness in CGI areas which demand seamless textures. There are plenty of automatic and A.I used methods to turn texture into seamless, many of them built-in free and commercial graphic tools. While we have uploaded the raw textures (not seamless). We also examine a few ways to make them seamless. Two common options are using inpainting of edge regions using models such as stable diffusion. This is done by tiling the textures leaving space between them for the model to inpaint and adding prompts such as "tileable texture". The code for this is available at this URL. This approach using stable diffusion 3 gives relatively unimpressive results. An alternative standard approach is simply cross-fading or blurring the opposite sides of the texture into each other creating a smooth transition. The code for this is available at this URL. However, this gives poor results when dealing with sharp textures and binary states such as black and white checkboards.

### A8.1. Seamless tiling using texture features topology

Ideally, we want the stitching of the tiles in seamless textures to be done using the features of the textures. The stitching of adjacent textures will vary between different textures but will be done using the texture native patterns, and not arbitrary methods like blurring. This will make the tile's boundaries more native to the image and harder to notice (Figure 5. right). To do this we consider the image as a topology meaning one property of the image is considered as a height map (for example pixel intensity).  When blending to opposite sides of textures we created a small overlap between them and identified using the topology which side is on top and which is bottom according to the height maps. Meaning in the overlap region pixels from the side that have a higher height will be shown, and the other pixels will be occluded. This allows us to stitch the image by only showing the top side in the stitch region, while the stitch pattern itself is derived from the image pattern (Figure 5.right). Finally, when moving across the stitch region we gradually erode the topology on one side and dilate it on the other to achieve a seamless transition. This blending method adapts to the image's native topology, minimizing visual seams. We supply both the raw textures (not seamless) and the seamless textures. The full code is available at this URL.

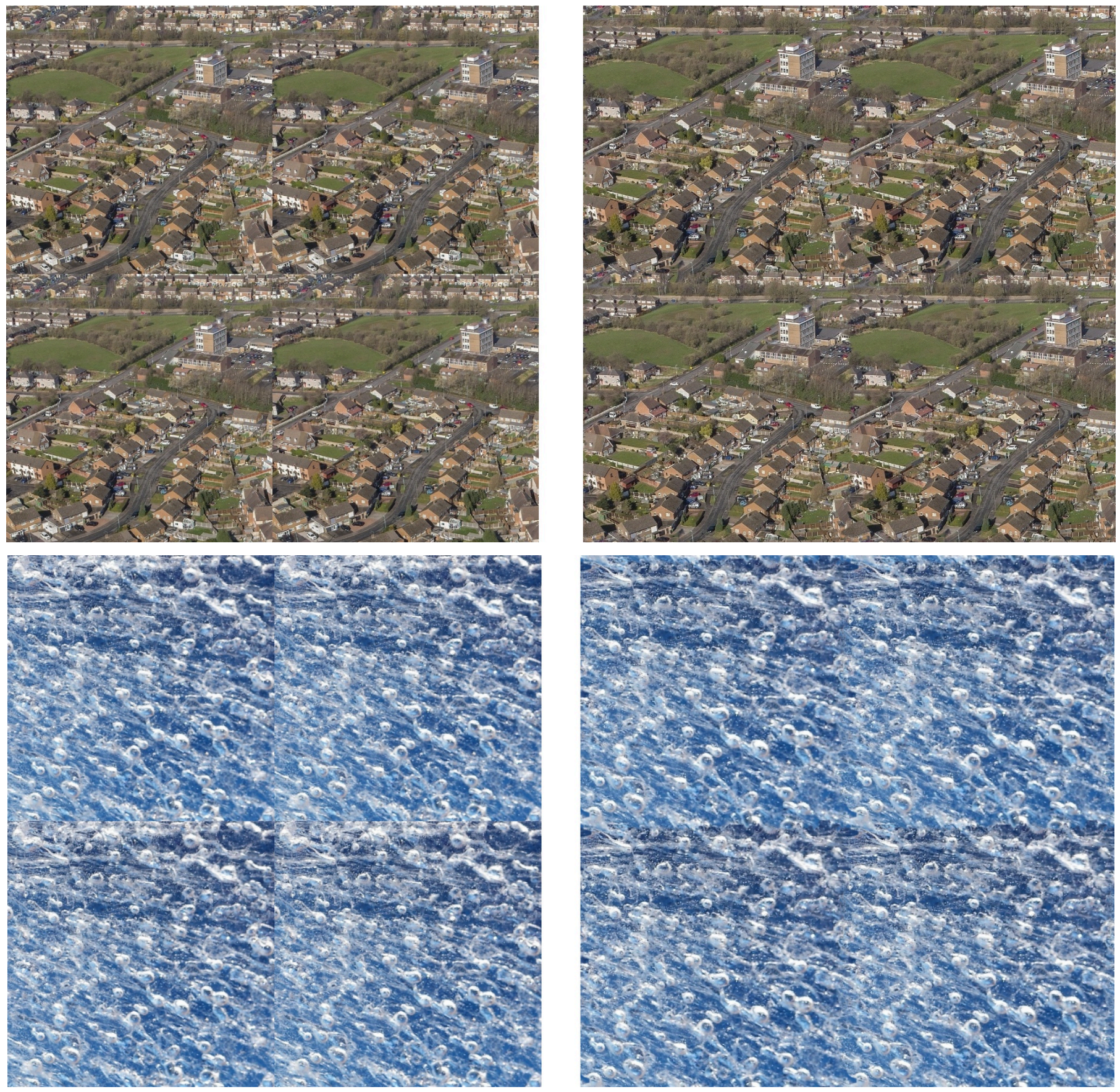

**Figure 5) Example of seamless tiling of images and textures using a topological approach. Left: Tiled grid of raw unmodified images. Right: Tiled images were modified using the topology approach (A8.1).**

## A10. Image sources and textures size

Another limitation of the Vastexture is the textures sizes which are mostly between 240-1000 pixels wide. This is due to the need for large images in order to extract large textures. For the image repositories we tested, finding textures that are larger than ¼ of the image size is rare and more than ½ is very rare. The segment of anything (SAM) repository contains images of sizes around 2000 pixels and enables extraction of textures between 512 -1000 pixels, about 100k such textures were extracted and converted to PBR (about 1 texture per 5 images). The open images repository has a very permissive license (Apache) but small images. Extracting textures of sizes >1000 pixels will demand a large repository of very large images >4000 pixels per dimension. We are not aware of any such large scale repository with free license.

## A11. Dataset source and license

The Vastexture repository available for downloads at this URLs: 1, 2, 3, 4
The code used to generate the repository available with CC0 license at 1, 2, 3, 4 .
The dataset license is CC0 but the license for each texture is derived from the source image license (open images, segment anything) .